%% file: aistats2019.tex
\documentclass[twoside]{article}

\usepackage[accepted]{aistats2019}

\usepackage[utf8]{inputenc} 
\usepackage[T1]{fontenc}    
\usepackage{url}            
\usepackage{booktabs}       
\usepackage{amsfonts}       
\usepackage{nicefrac}       
\usepackage{microtype}      
\usepackage{mathtools}
\usepackage{hyperref}      
\usepackage{amsmath}
\usepackage{amsthm}
\usepackage{amsfonts}
\usepackage{amssymb}
\usepackage{graphicx}
\usepackage{booktabs}
\usepackage{placeins}
\usepackage{multirow}
\usepackage[ruled, linesnumbered]{algorithm2e}
\usepackage{epstopdf}
\usepackage{color}
\usepackage{subcaption}
\usepackage[T1]{fontenc}
\usepackage{lmodern}
\usepackage{amsmath,amsfonts,bm}
\usepackage{balance}
\usepackage{enumitem}
\usepackage{xcolor}
\usepackage{balance}
\usepackage{wrapfig}
\usepackage{subcaption}
\usepackage{enumitem}

\usepackage[round]{natbib}

\runningtitle{Confidence-based Graph Convolutional Networks for Semi-Supervised Learning}

\begin{document}

%

\def\vec#1{\mbox{\bf #1}}
\def\mat#1{\mbox{\bf #1}}
%

\newcommand{\customfootnotetext}[2]{{
		\renewcommand{\thefootnote}{#1}
		\footnotetext[0]{#2}}}

\twocolumn[

\aistatstitle{Confidence-based Graph Convolutional Networks \\ for Semi-Supervised Learning}

\aistatsauthor{
	Shikhar Vashishth$^{\star}$ \quad Prateek Yadav$^{\star}$ \quad Manik Bhandari$^{\star}$ \quad \textbf{Partha Talukdar}\\
    Indian Institute of Science \\
	{\tt \small \{shikhar,prateekyadav,manikb,ppt\}@iisc.ac.in} \\

}

\aistatsaddress{} ]

\customfootnotetext{$^{\star}$}{Equal contribution}

\input{defs}

\begin{abstract}
	Predicting properties of nodes in a graph is an important problem with applications in a variety of domains. Graph-based Semi Supervised Learning (SSL) methods aim to address this problem by labeling a small subset of the nodes as seeds and then utilizing the graph structure to predict label scores for the rest of the nodes in the graph. Recently, Graph Convolutional Networks (GCNs) have achieved impressive performance on the graph-based SSL task. In addition to label scores, it is also desirable to have confidence scores associated with them. Unfortunately, confidence estimation in the context of GCN has not been previously explored. We fill this important gap in this paper and propose \method{}, which estimates labels scores along with their confidences jointly in GCN-based setting. \method{} uses these estimated confidences to determine the influence of one node on another during neighborhood aggregation, thereby acquiring \emph{anisotropic}\footnote{anisotropic (adjective): varying in magnitude according to the direction of measurement (Oxford English Dictionary)} capabilities. Through extensive analysis and experiments on standard benchmarks, we find that \method{} is able to outperform state-of-the-art baselines. We have made \method{}'s source code available to encourage reproducible research.
\end{abstract}

\input{sections/intro}
\input{sections/related_work}
\input{sections/back}
\input{sections/overview}
\input{sections/details}

\input{sections/experiments}

\input{sections/results}
\input{sections/conclusion}

\bibliographystyle{apalike}

\bibliography{references}

\end{document}

%% file: defs.tex
\newcommand{\refalg}[1]{Algorithm \ref{#1}}
\newcommand{\refeqn}[1]{Equation \ref{#1}}
\newcommand{\reffig}[1]{Figure \ref{#1}}
\newcommand{\reftbl}[1]{Table \ref{#1}}
\newcommand{\refsec}[1]{Section \ref{#1}}

\newcommand{\reminder}[1]{\textcolor{red}{[[ #1 ]]}\typeout{#1}}
\newcommand{\reminderR}[1]{\textcolor{gray}{[[ #1 ]]}\typeout{#1}}

\newcommand{\add}[1]{\textcolor{red}{#1}\typeout{#1}}
\newcommand{\remove}[1]{\sout{#1}\typeout{#1}}

\newcommand{\m}[1]{\mathcal{#1}}
\newcommand{\method}{ConfGCN}

\newtheorem{theorem}{Theorem}[section]
\newtheorem{claim}[theorem]{Claim}

\newcommand{\tensor}{\mathcal{X}}
\newcommand{\Real}{\mathbb{R}}

\newcommand{\tuples}{\mathbb{T}}

\newcommand\norm[1]{\left\lVert#1\right\rVert}

\newcommand{\note}[1]{\textcolor{blue}{#1}}

\newcommand*{\Scale}[2][4]{\scalebox{#1}{$#2$}}%
\newcommand*{\Resize}[2]{\resizebox{#1}{!}{$#2$}}%

\def\mat#1{\mbox{\bf #1}}

%% file: sections/intro.tex
\section{Introduction}
\label{sec:intro}

Graphs are all around us, ranging from citation and social networks to knowledge graphs. Predicting properties of nodes in such graphs is often desirable. For example, given a citation network, we may want to predict the research area of an author. Making such predictions, especially in the semi-supervised setting, has been the focus of graph-based semi-supervised learning (SSL) \citep{subramanya2014graph}. In graph-based SSL, a small set of nodes are initially labeled. Starting with such supervision and while utilizing the rest of the graph structure, the initially unlabeled nodes are labeled. Conventionally, the graph structure has been  incorporated as an explicit regularizer which enforces a  smoothness constraint on the labels estimated on nodes  \citep{Zhu2003semi,Belkin2006manifold,Weston2008deep}. 
Recently proposed Graph Convolutional Networks (GCN) \citep{Defferrard2016,Kipf2016} provide a framework to apply deep neural networks to graph-structured data. GCNs have been employed successfully for improving performance on tasks such as  semantic role labeling \citep{gcn_srl}, machine translation \citep{gcn_mt}, relation extraction \citep{reside,gcn_re}, document dating \citep{neuraldater}, shape segmentation \citep{yi2016syncspeccnn}, and action recognition \citep{huang2017deep}. GCN formulations for graph-based SSL have also attained state-of-the-art performance  \citep{Kipf2016,Liao2018graph,graph_attention_network}. In this paper, we also focus on the task of graph-based SSL using GCNs. 

GCN iteratively estimates embedding of nodes in the graph by aggregating embeddings of neighborhood nodes, while backpropagating errors from a target loss function. 
Finally, the learned node embeddings are used to estimate label scores on the nodes. In addition to the label scores, it is desirable to also have confidence estimates associated with them. Such confidence scores may be used to determine how much to trust the label scores estimated on a given node. While methods to estimate label score confidence in non-deep graph-based SSL has been previously proposed \citep{Orbach2012}, confidence-based GCN is still unexplored. 

\begin{figure*}[t]
	\centering
	\includegraphics[scale=0.55]{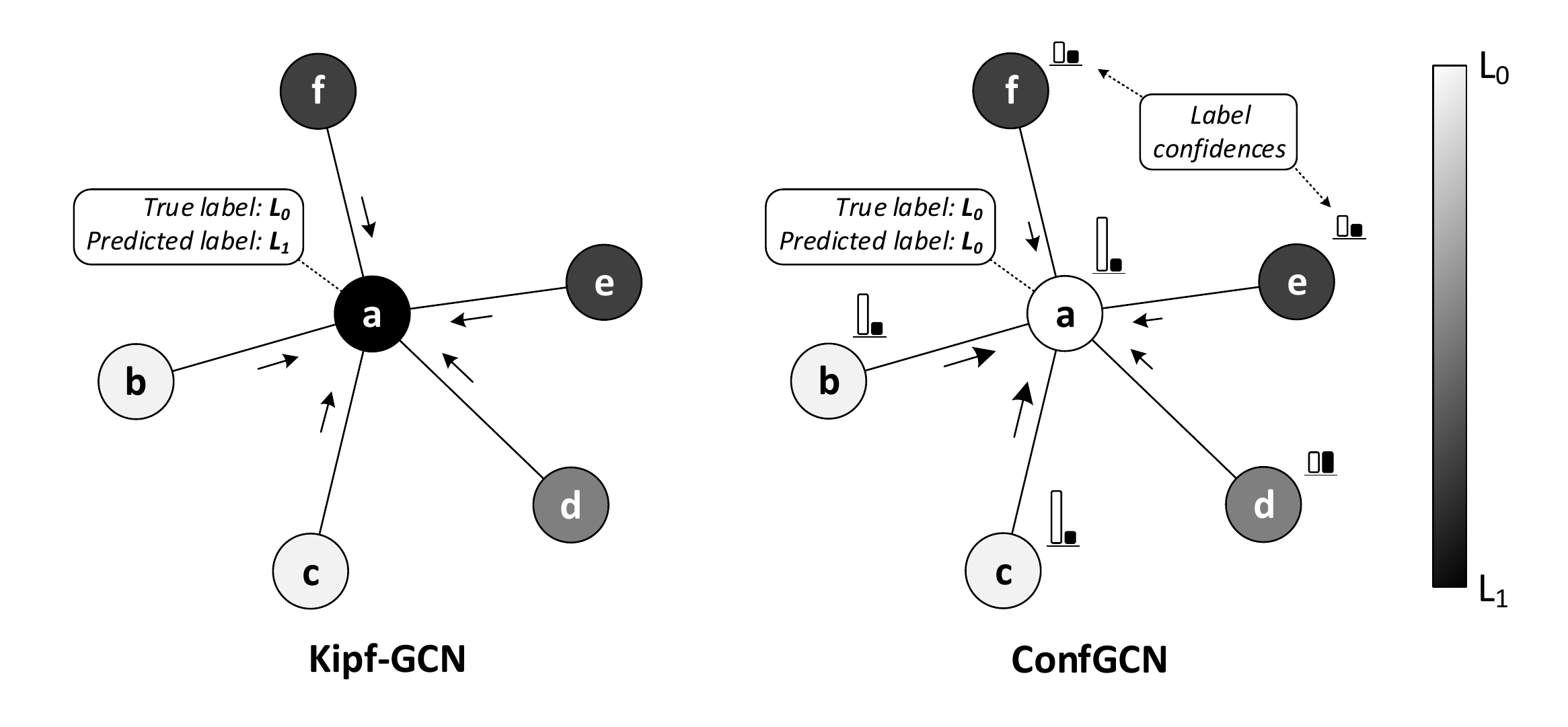}
	\caption{\label{fig:motivation}Label prediction on node $a$ by Kipf-GCN and \method{} (this paper). $L_0$ is $a$'s true label. Shade intensity of a node reflects the estimated score of label $L_1$ assigned to that node. 
	Since Kipf-GCN is not capable of estimating influence of one node on another, it is misled by the dominant label $L_1$ in node $a$'s neighborhood and thereby making the wrong assignment. \method{}, on the other hand, estimates confidences (shown by bars) over the label scores, and uses them to increase influence of nodes $b$ and $c$ to estimate the right label on $a$. Please see \refsec{sec:intro} for details. 
	}
\end{figure*}

In order to fill this important gap, we propose \method{}, a GCN framework for graph-based SSL. \method{} jointly estimates label scores on nodes, along with confidences over them.  One of the added benefits of confidence over node's label scores is that they may be used to subdue irrelevant nodes in a node's neighborhood, thereby controlling the number of effective neighbors for each node. In other words, this enables \emph{anisotropic} behavior in GCNs. Let us explain this through the example shown in \reffig{fig:motivation}. In this figure, while node $a$ has true label $L_0$ (white), it is incorrectly classified as $L_1$ (black) by Kipf-GCN \citep{Kipf2016}\footnote{In this paper, unless otherwise stated, we refer to Kipf-GCN whenever we mention GCN.}. This is because Kipf-GCN suffers from limitations of its  neighborhood aggregation scheme \citep{Xu2018}. For example, Kipf-GCN has no constraints on the number of nodes that can influence the representation of a given target node. In a $k$-layer Kipf-GCN model, each node is influenced by all the nodes in its $k$-hop neighborhood. 
However, in real world graphs, nodes are often present in  \emph{heterogeneous} neighborhoods, i.e., a node is often surrounded by nodes of other labels. For example, in \reffig{fig:motivation}, node $a$ is surrounded by three nodes ($d$, $e$, and $f$) which are predominantly labeled $L_1$, while two nodes ($b$ and $c$) are labeled $L_0$. Please note that all of these are estimated label scores during GCN learning. In this case, it is desirable that node $a$ is more influenced by nodes $b$ and $c$ than the other three nodes. However, since Kipf-GCN doesn't discriminate among the neighboring nodes, it is swayed by the majority and thereby estimating the wrong label $L_1$ for node $a$. 

\method{} is able to overcome this problem by estimating confidences on each node's label scores. In \reffig{fig:motivation}, such estimated confidences are shown by bars, with white and black bars denoting confidences in scores of labels $L_0$ and $L_1$, respectively. \method{} uses these label confidences to subdue nodes $d, e, f$ since they have low confidence for their label $L_1$ (shorter black bars), whereas nodes $b$ and $c$ are highly confident about their labels being $L_0$ (taller white bars). This leads to higher influence of $b$ and $c$ during aggregation, and thereby  \method{} correctly predicting the true label of node $a$ as $L_0$ with high confidence. This clearly demonstrates the benefit of label confidences and their utility in estimating node influences. Graph Attention Networks (GAT) \citep{graph_attention_network}, a recently proposed method also provides a mechanism to estimate  influences by allowing nodes to attend to their neighborhood. However, as we shall see in \refsec{sec:experiments}, \method{}, through its use of label confidences, is considerably more effective.

Our contributions in this paper are as follows.

\begin{itemize}
	\item We propose \method{}, a Graph Convolutional Network (GCN) framework for semi-supervised learning which models label distribution and their confidences for each node in the graph. To the best of our knowledge, this is the first confidence-enabled formulation of GCNs.
	\item \method{} utilize label confidences to estimate influence of one node on another in a label-specific manner during neighborhood aggregation of GCN learning.
	\item Through extensive evaluation on multiple real-world datasets, we demonstrate \method{} effectiveness over state-of-the-art baselines.
\end{itemize}

\method{}'s source code and datasets used in the paper are available at \url{http://github.com/malllabiisc/ConfGCN}.

%% file: sections/related_work.tex
\section{Related Work}
\label{sec:related_work}
\textbf{Semi-Supervised learning (SSL) on graphs:} SSL on graphs is the problem of classifying nodes in a graph, where labels are available only for a small fraction of nodes. Conventionally, the graph structure is imposed by adding an explicit graph-based regularization term in the loss function \citep{Zhu2003semi,Weston2008deep,Belkin2006manifold}.
Recently, implicit graph regularization via learned node representation has proven to be more effective. 
This can be done either sequentially or in an end to end fashion. Methods like DeepWalk \citep{Perozzi2014deepWalk}, node2vec \citep{Grover2016}, and LINE \citep{Tang2015} first learn graph representations via sampled random walk on the graph or breadth first search traversal and then use the learned representation for node classification. On the contrary, Planetoid \citep{Yang2016revisiting} learns node embedding by jointly predicting the class labels and the neighborhood context in the graph.
Recently, \cite{Kipf2016} employs Graph Convolutional Networks (GCNs) to learn node representations.

\textbf{Graph Convolutional Networks (GCNs):}
The generalization of Convolutional Neural Networks to non-euclidean domains is proposed by \cite{Bruna2013} which formulates the spectral and spatial construction of GCNs. This is later improved through an efficient localized filter approximation \citep{Defferrard2016}. \cite{Kipf2016} provide a first-order formulation of GCNs and show its effectiveness for SSL on graphs. \cite{gcn_srl} propose GCNs for directed graphs and provide a mechanism for edge-wise gating to discard noisy edges during aggregation. This is further improved by \cite{graph_attention_network} which allows nodes to attend to their neighboring nodes, implicitly providing different weights to different nodes. \cite{Liao2018graph} propose Graph Partition Neural Network (GPNN), an extension of GCNs to learn node representations on large graphs. GPNN first partitions the graph into subgraphs and then alternates between locally and globally propagating information across subgraphs. Recently, Lovasz Convolutional Networks \cite{lovasz_paper} is proposed for incorporating global graph properties in GCNs. An extensive survey of GCNs and their applications can be found in \cite{Bronstein2017}.

\textbf{Confidence Based Methods:} The natural idea of incorporating confidence in predictions has been explored by \cite{Li2006} for the task of active learning. \cite{Lei2014} proposes a confidence based framework for classification problems, where the classifier consists of two regions in the predictor space, one for confident classifications and other for ambiguous ones. In representation learning, uncertainty (inverse of confidence) is first utilized for word embeddings by \cite{Vilnis2014}. 
\cite{Athiwaratkun2018} further extend this idea to learn hierarchical word representation through encapsulation of probability distributions. 
\cite{Orbach2012} propose TACO (Transduction Algorithm with COnfidence), the first graph based method which learns
label distribution along with its uncertainty for semi-supervised node classification. 
\cite{graph2gauss} embeds graph nodes as Gaussian distribution using ranking based framework which allows to capture uncertainty of representation. They update node embeddings to maintain neighborhood ordering, i.e. 1-hop neighbors are more similar to 2-hop neighbors and so on. Gaussian embeddings have been used for collaborative filtering \citep{DosSantos2017} and topic modelling \citep{Das2015} as well.

%% file: sections/back.tex

\section{Notation \& Problem Statement} 
\label{sec:notation}

Let $\mathcal{G} = (\mathcal{V},\mathcal{E},\mathcal{X})$ be an undirected graph, where $\mathcal{V} = \mathcal{V}_l \cup \mathcal{V}_u$ is the union of labeled ($\mathcal{V}_l$) and unlabeled ($\mathcal{V}_u$) nodes in the graph with cardinalities $n_l $ and $n_u$, $\mathcal{E}$ is the set of edges and $\mathcal{X} \in \mathbb{R}^{(n_l + n_u) \times d}$ is the input node features.
The actual label of a node $v$ is denoted by a one-hot vector $Y_v \in \mathbb{R}^{m}$, where $m$ is the number of classes. Given $\mathcal{G}$ and seed labels $Y \in \mathbb{R}^{n_l \times m}$, the goal is to predict the labels of the unlabeled nodes. 
To incorporate confidence, we additionally estimate label distribution $\bm{\mu}_v \in \mathbb{R}^{m}$ and a diagonal co-variance matrix $\bm{\Sigma}_v \in \mathbb{R}^{m \times m},~\forall v \in \mathcal{V}$. Here, $\bm{\mu}_{v,i}$ denotes the score of label $i$ on node $v$, while $(\bm{\Sigma}_v)_{ii}$ denotes the variance in the estimation of $\bm{\mu}_{v,i}$. In other words, $(\bm{\Sigma}_{v}^{-1})_{ii}$ is \method{}'s confidence in $\bm{\mu}_{v,i}$.

\section{Background: Graph Convolutional Networks}
\label{sec:background}

In this section, we give a brief overview of Graph Convolutional Networks (GCNs) for undirected graphs as proposed by \cite{Kipf2016}. Given a graph $G = (\m{V}, \m{E}, \m{X})$ as defined \refsec{sec:notation}, the node representation after a single layer of GCN can be defined as
\begin{equation}
\bm{H} = f((\bm{\tilde{D}}^{-\frac{1}{2}} (\bm{A}+ \bm{I}) \bm{\tilde{D}}^{-\frac{1}{2}})\m{X} \bm{W})
\label{eqn:kipf_basic}
\end{equation}
where, $\bm{W} \in \mathbb{R}^{d \times d}$ denotes the model parameters, $\bm{A}$ is the adjacency matrix and $\bm{\tilde{D}}_{ii} = \sum_{j}(\bm{A}+\bm{I})_{ij}$. $f$ is any activation function, we have used ReLU, $f(x) = \max(0, x)$ in this paper. \refeqn{eqn:kipf_basic} can also be written as
\begin{equation}
\bm{h}_{v} = f \left(\sum_{u \in \mathcal{N}(v)}\bm{W} \bm{h}_{u} + \bm{b} \right) ,~~~\forall v \in \mathcal{V}.
\label{eqn:gcn_undirected}
\end{equation}

Here, $\bm{b} \in \mathbb{R}^d$ denotes bias, $\m{N}(v)$ corresponds to immediate neighbors of $v$ in graph $\m{G}$ including $v$ itself and $\bm{h}_{v}$ is the obtained representation of node $v$.

For capturing multi-hop dependencies between nodes, multiple GCN layers can be stacked on top of one another. The representation of node $v$ after $k^{th}$ layer of GCN is given as
\begin{equation}
\bm{h}^{k+1}_{v} = f\left(\sum_{u \in \m{N}(v)}\left(\bm{W}^{k} \bm{h}^{k}_{u} + \bm{b}^{k}\right)\right), \forall v \in \m{V} .
\label{eqn:gcn_undir_main}
\end{equation}

where,  $\bm{W}^{k}, \bm{b}^{k}$ denote the layer specific parameters of GCN.

%% file: sections/details.tex
\section{Confidence Based Graph Convolution (\method{})}
\label{sec:details}

Following \citep{Orbach2012}, \method{} 
%
uses co-variance matrix based symmetric Mahalanobis distance for defining distance between two nodes in the graph. Formally, for any two given nodes $u$ and $v$, with label distributions $\bm{\mu}_u$ and $\bm{\mu}_v$  and co-variance matrices $\bm{\Sigma}_u$ and $\bm{\Sigma}_v$, distance between them is defined as follows.
\[
d_M(u,v) = (\bm{\mu}_u-\bm{\mu}_v)^T(\bm{\Sigma}_u^{-1}+\bm{\Sigma}_v^{-1})(\bm{\mu}_u-\bm{\mu}_v). 
\]
Characteristic of the above distance metric is that if either of $\bm{\Sigma}_u$ or $\bm{\Sigma}_v$ has large eigenvalues, then the distance will be low irrespective of the closeness of $\bm{\mu}_u$ and $\bm{\mu}_v$. On the other hand, if $\bm{\Sigma}_u$ and $\bm{\Sigma}_v$ both have low eigenvalues,  then it requires $\bm{\mu}_u$ and $\bm{\mu}_v$ to be close for their distance to be low. Given the above properties, we define $r_{uv}$, the influence score of node $u$ on its neighboring node $v$ during GCN aggregation, as follows.
\[
r_{uv} = \frac{1}{d_M(u,v)} .
\]
This influence score gives more relevance to neighboring nodes with highly confident similar label, while reducing importance of nodes with low confident label scores. This results in \method{} acquiring anisotropic capability during neighborhood aggregation. For a node $v$, \method{}'s  equation for updating embedding at the $k$-th layer is thus defined as follows.

\begin{equation}
\bm{h}^{k+1}_{v} = f\left(\sum_{u \in \m{N}(v)}  r_{uv} \times \left(\bm{W}^{k} \bm{h}^{k}_{u} + \bm{b}^{k}\right)\right), \forall v \in \m{V} .
\end{equation}

%
The final node representation obtained from \method{} is used for predicting labels of the nodes in the graph as follows.

\[
\bm{\hat{Y}}_v = \mathrm{softmax}(\bm{W}^K \bm{h}^K_v + \bm{b}^K), ~\forall v \in \m{V}
\]

where, $K$ denotes the number of \method{}'s layers. Finally, in order to learn label scores $\{\bm{\mu}_{v}\}$ and co-variance matrices $\{\bm{\Sigma}_v\}$ jointly with other parameters $\{\bm{W}^k, \bm{b}^k\}$, following \cite{Orbach2012}, we include the following two terms in \method{}'s objective function.

%

\begin{table*}[!tbh]
	\centering
	\begin{tabular}{lcccccc}
		\toprule
		Dataset  & Nodes & Edges & Classes & Features & Label Mismatch & $\frac{|\mathcal{V}_l|}{|\mathcal{V}|}$ \\
		\midrule
		Cora 			  		& 2,708 		 & 5,429 		& 7  	 & 1,433 		&  0.002 & 0.052\\
		Cora-ML				 & 2,995		  & 8,416 		  & 7		& 2,879		  & 0.018 & 0.166\\
		Citeseer 		 		& 3,327 		& 4,372 		& 6 	& 3,703 	&	0.003 & 0.036 \\
		Pubmed 		  		 & 19,717 		  & 44,338 		& 3  	 & 500   	&	0.0  & 0.003\\
		\bottomrule
	\end{tabular}
	\caption{\label{tbl:dataset_statistics}Details of the datasets used in the paper. Please refer \refsec{sec:datasets} for more details.}
\end{table*} 

For enforcing neighboring nodes to be close to each other, we include $L_{\text{smooth}}$ defined as
\vspace{-2mm}

\[L_{\text{smooth}} = \sum_{(u,v) \in \m{E}} (\bm{\mu}_u- \bm{\mu}_v)^T(\bm{\Sigma}^{-1}_u + \bm{\Sigma}^{-1}_v)(\bm{\mu}_u - \bm{\mu}_v). \]

To impose the desirable property that the label distribution of nodes in $\m{V}_l$ should be close to their input label distribution, we incorporate $L_{\text{label}}$ defined as 
\vspace{-3mm}

\[L_{\text{label}}  = \sum_{v \in \m{V}_l} (\bm{\mu}_v - \bm{Y}_v)^T(\bm{\Sigma}^{-1}_v + \dfrac{1}{\gamma}\bm{I})(\bm{\mu}_v - \bm{Y}_v)  .\]

Here, for input labels, we assume a fixed uncertainty of $\frac{1}{\gamma}\bm{I} \in \mathbb{R}^{L \times L}$, where $\gamma > 0$. 
We also include the following regularization term, $L_{\text{reg}}$ to constraint the co-variance matrix to be finite and positive. 

\[L_{\text{reg}}  = \sum_{v \in \m{V}}\mathrm{Tr} \ \mathrm{max}(-\bm{\Sigma}_v, 0)  ,\]

This regularization term enforces soft positivity constraint on co-variance matrix. 
Additionally in \method{}, we include the $L_{\text{const}}$ in the objective, to push the label distribution ($\bm{\mu}$) close to the final model prediction ($\bm{\hat{Y}}$).

\[L_{\text{const}}  = \sum_{v \in \m{V}} (\bm{\mu}_v - \hat{\bm{Y}}_v)^T(\bm{\mu}_v - \hat{\bm{Y}}_v) .\]

Finally, we include the standard cross-entropy loss for semi-supervised multi-class classification over all the labeled nodes ($\mathcal{V}_l$).
\vspace{-2mm}

\[L_{\text{cross}} = - \sum_{v \in \mathcal{V}_l} \sum_{j=1}^{m} \bm{Y}_{vj} \log(\bm{\hat{Y}}_{vj})  .\]

The final objective for optimization is the linear combination of the above defined terms. 
\vspace{-2mm}


\begin{equation*} \label{eqn:main_obj}
\small
\begin{aligned}
L(\theta) = 
& \hspace{0 mm} - \sum_{i \in \mathcal{V}_l} \sum_{j=1}^{L} \bm{Y}_{ij} \log( \bm{\hat{Y}}_{ij}) \\    
& + \hspace{1 mm} \lambda_{\text{1}} \sum_{(u,v) \in \m{E}} (\bm{\mu}_u- \bm{\mu}_v)^T(\bm{\Sigma}^{-1}_u + \bm{\Sigma}^{-1}_v)(\bm{\mu}_u - \bm{\mu}_v) \\
& + \hspace{1 mm} \lambda_{\text{2}} \hspace{1.5 mm} \sum_{u \in \m{V}_l} (\bm{\mu}_u - \bm{Y}_u)^T(\bm{\Sigma}^{-1}_u + \dfrac{1}{\gamma}\bm{I})(\bm{\mu}_u - \bm{Y}_u)  \\
& + \hspace{1 mm} \lambda_{\text{3}} \hspace{2 mm} \sum_{v \in \m{V}} (\bm{\mu}_u - \hat{\bm{Y}}_u)^T(\bm{\mu}_u - \hat{\bm{Y}}_u)\\
& + \hspace{1 mm} \lambda_{\text{4}} \hspace{2 mm} \sum_{v \in \m{V}}\mathrm{Tr} \ \mathrm{max}(-\bm{\Sigma}_v, 0)  \\
\end{aligned}
\end{equation*}

where, $\theta = \{\bm{W}^k, \bm{b}^k, \bm{\mu}_v, \bm{\Sigma}_v\}$ and $\lambda_i \in \mathbb{R}$, are the weights of the terms in the objective. We optimize $L(\theta)$ using stochastic gradient descent. We hypothesize that all the terms help in improving \method{}'s performance and we validate this in \refsec{sec:ablation_results}.

%% file: sections/experiments.tex
\section{Experiments}
\label{sec:experiments}
\subsection{Datasets}
\label{sec:datasets}

For evaluating the effectiveness of \method{}, we evaluate on several semi-supervised classification benchmarks. Following the experimental setup of \citep{Kipf2016,Liao2018graph}, we evaluate on Cora, Citeseer, and Pubmed datasets \citep{sen:aimag08}. The dataset statistics is summarized in \reftbl{tbl:dataset_statistics}. Label mismatch denotes the fraction of edges between nodes with different labels in the training data. The benchmark datasets commonly used for semi-supervised classification task have substantially low label mismatch rate. In order to examine models on datasets with more heterogeneous neighborhoods, 
we also evaluate on Cora-ML dataset \citep{graph2gauss}. 


All the four datasets are citation networks, where each document is represented using bag-of-words features  in the graph with undirected citation links between documents. The goal is to classify documents into one of the predefined classes. We use the data splits used by \citep{Yang2016revisiting} and follow similar setup for Cora-ML dataset. Following \citep{Kipf2016}, additional 500 labeled nodes are used for hyperparameter tuning.


\begin{table*}[t]
	\centering
	\begin{tabular}{lcccc}
		\toprule
		
		Method & Citeseer & Cora & Pubmed & Cora ML\\
		\midrule
		
		LP 	\citep{Zhu2003semi}	& 45.3 		& 68.0 		& 63.0 		& -\\
		ManiReg \citep{Belkin2006manifold}	& 60.1 		& 59.5 		& 70.7 		& -\\
		SemiEmb \citep{Weston2008deep}		& 59.6 		& 59.0 		& 71.1 		& -\\
		Feat 	\citep{Yang2016revisiting}	& 57.2 		& 57.4 		& 69.8 		& -\\
		DeepWalk \citep{Perozzi2014deepWalk}	& 43.2 		& 67.2 		& 65.3 		& -\\
		GGNN	\citep{Li2015gated}	& 68.1 		& 77.9 		& 77.2 		& -\\
		Planetoid \citep{Yang2016revisiting}	& 64.9 		& 75.7 		& 75.7 		& -\\
		Kipf-GCN \citep{Kipf2016}			& 69.4 $\pm$ 0.4 	& 80.9 $\pm$ 0.4 	& 76.8 $\pm$ 0.2 	& 85.7 $\pm$ 0.3\\
		G-GCN \citep{gcn_srl}				& 69.6 $\pm$ 0.5 		& 81.2 $\pm$ 0.4 		& 77.0 $\pm$ 0.3 		& 86.0 $\pm$ 0.2 \\
		GPNN \citep{Liao2018graph} 			& 68.1 $\pm$ 1.8		& 79.0 $\pm$ 1.7 		&  73.6 $\pm$ 0.5		& 69.4 $\pm$ 2.3 \\
		GAT \citep{graph_attention_network} & 72.5 $\pm$ 0.7 		& \textbf{83.0 $\pm$ 0.7} 		& 79.0 $\pm$ 0.3 		& 83.0 $\pm$ 0.8\\
		\midrule
		ConfGCN (this paper)	 & \textbf{72.7 $\pm$ 0.8} & 82.0 $\pm$ 0.3 & \textbf{79.5 $\pm$ 0.5} & \textbf{86.5 $\pm$ 0.3}\\
		\bottomrule
	\end{tabular}
	\caption{\label{tbl:qualitative_results} Performance comparison of several methods for semi-supervised node classification on multiple benchmark datasets. \method{} performs consistently better across all the datasets. Baseline method performances on Citeseer, Cora and Pubmed datasets are taken from \cite{Liao2018graph,graph_attention_network}. We consider only the top performing baseline methods on these datasets for evaluation on the Cora-ML dataset. Please refer \refsec{sec:quantitative_results} for details.}
	
\end{table*}

\textbf{Hyperparameters:} We use the same data splits as described in \citep{Yang2016revisiting}, with a test set of 1000 labeled nodes for testing the prediction accuracy of \method{} and a validation set of 500 labeled nodes for optimizing the hyperparameters. The ranges of hyperparameters were adapted from previous literature \citep{Orbach2012,Kipf2016}. The model is trained using Adam \citep{kingma2014adam} with a learning rate of 0.01. The weight matrices along with ${\bf \mu}$  are initialized using Xavier initialization \citep{glorot2010understanding} and ${\bf \Sigma}$ matrix is initialized with identity. To avoid numerical instability we model ${\bf \Sigma^{-1}}$ directly and compute ${\bf \Sigma}$ wherever required. Following \cite{Kipf2016}, we use two layers of GCN ($K$) for all the experiments in this paper.

\subsection{Baselines}
\label{sec:baselines}
For evaluating \method{}, we compare against the following baselines:
\begin{itemize}[leftmargin=*]
	\item \textbf{Feat} \citep{Yang2016revisiting} takes only node features as input and ignores the graph structure.
	\item \textbf{ManiReg} \citep{Belkin2006manifold} is a framework for providing data-dependent geometric regularization.
	\item \textbf{SemiEmb} \citep{Weston2008deep} augments deep architectures with semi-supervised regularizers to improve their training.
	\item \textbf{LP} \citep{Zhu2003semi} is an iterative iterative label propagation algorithm which propagates a nodes labels to its neighboring unlabeled nodes according to their proximity.
	\item \textbf{DeepWalk} \citep{Perozzi2014deepWalk} learns node features by treating random walks in a graph as the equivalent of sentences.
	\item \textbf{Planetoid} \citep{Yang2016revisiting} provides a transductive and inductive framework for  jointly predicting class label and neighborhood context of a node in the graph. 
	\item \textbf{GCN} \citep{Kipf2016} is a variant of convolutional neural networks used for semi-supervised learning on graph-structured data.
	\item \textbf{G-GCN} \citep{gcn_srl} is a variant of GCN with edge-wise gating to discard noisy edges during aggregation.
	\item \textbf{GGNN} \citep{Li2015gated} is a generalization of RNN framework which can be used for graph-structured data.
	\item \textbf{GPNN} \citep{Liao2018graph} is a graph partition based algorithm which propagates information after partitioning large graphs into smaller subgraphs.
	\item \textbf{GAT} \citep{graph_attention_network} is a graph attention based method which provides different weights to different nodes by allowing nodes to attend to their neighborhood.
\end{itemize}



%% file: sections/results.tex
\begin{figure*}[t!]
	\begin{minipage}{0.5\linewidth}
		\centering
		\includegraphics[width=\linewidth]{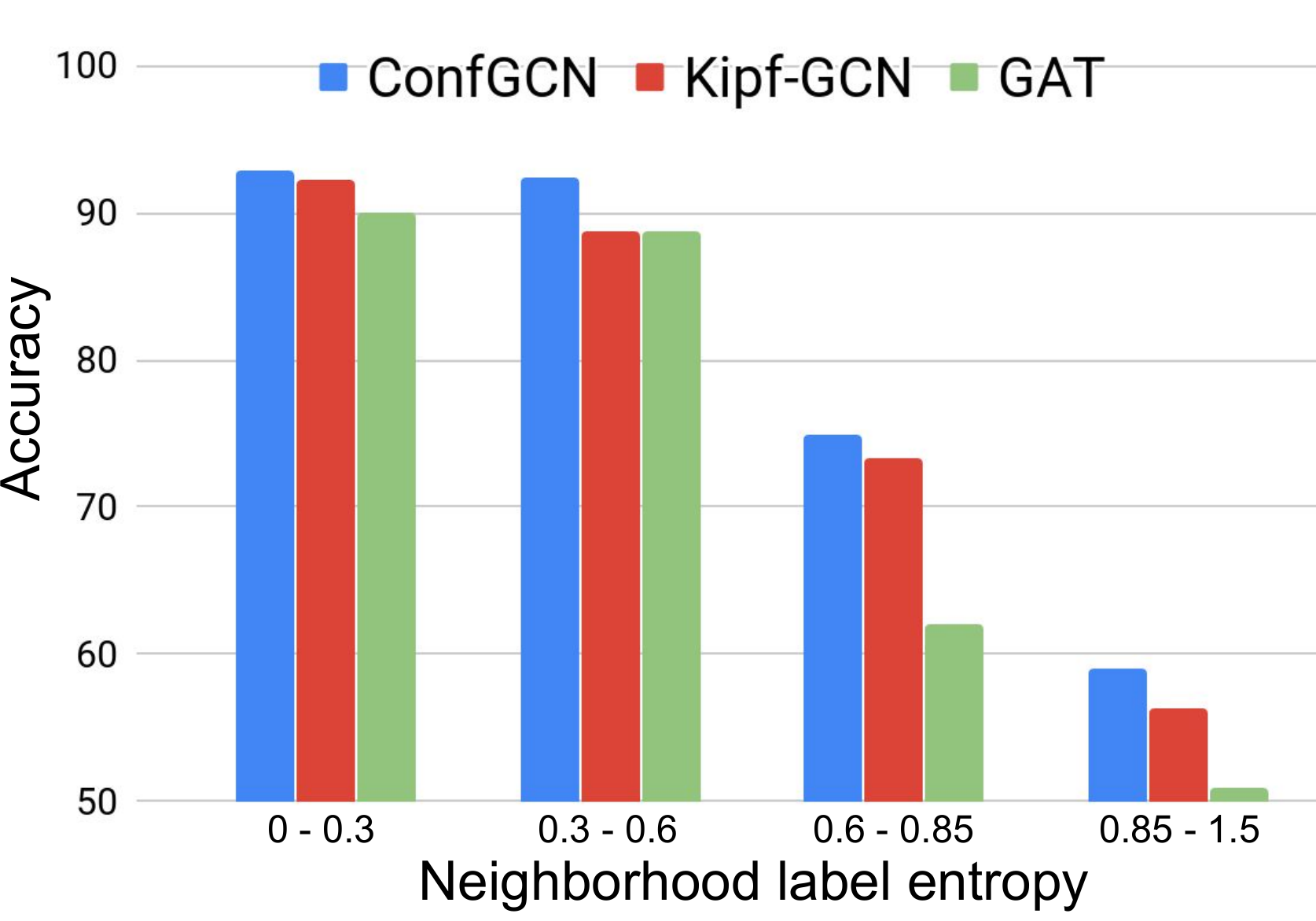}
		\subcaption{\label{fig:entropy_plot}}
	\end{minipage} 
	\hfill
	\begin{minipage}{0.5\linewidth}
		\centering
		\includegraphics[width=\linewidth]{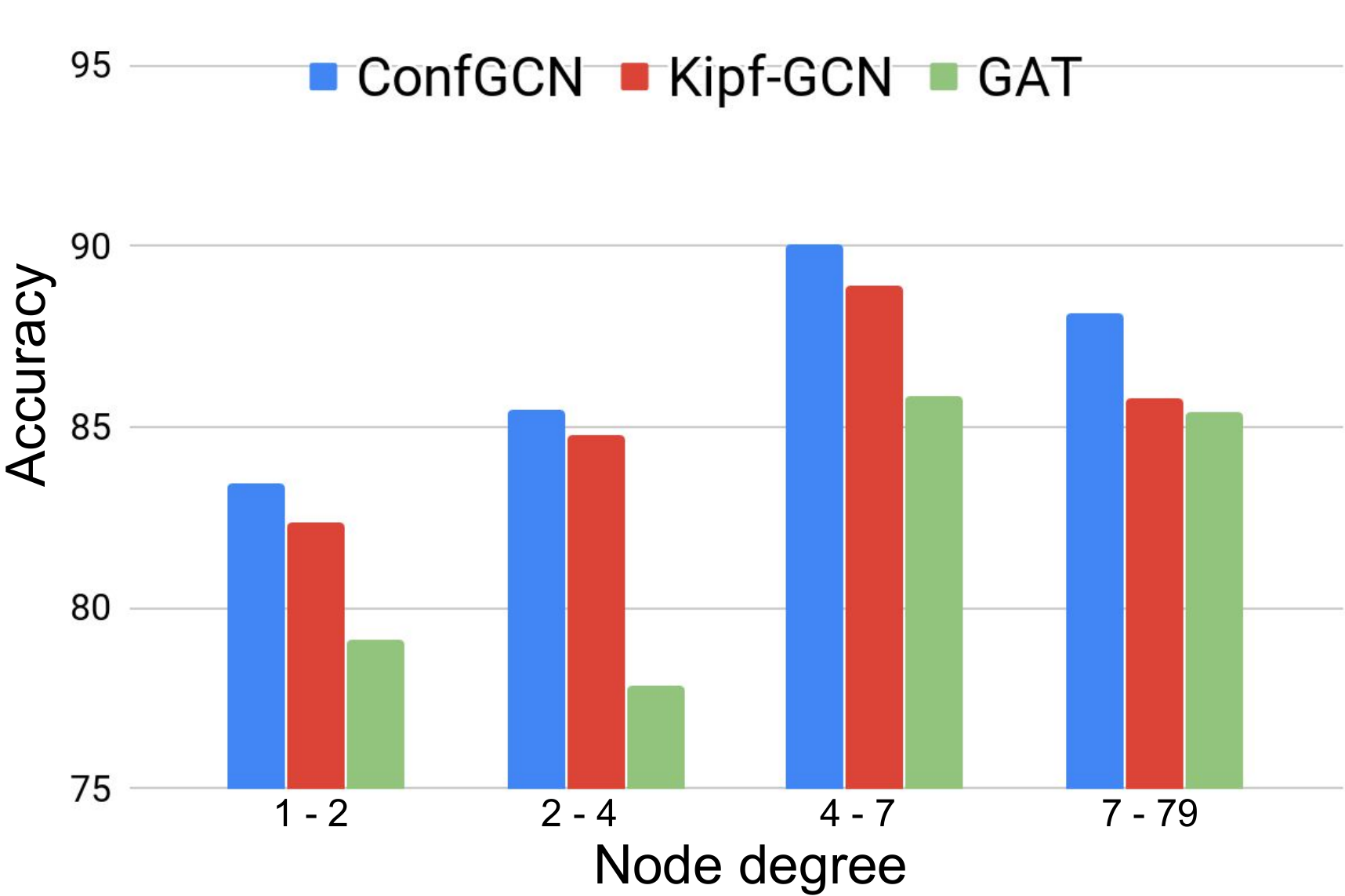}
		\subcaption{\label{fig:degree_plot}}
	\end{minipage}
	\caption{Plots of node classification accuracy vs. (a) neighborhood label entropy and (b) node degree. On $x$-axis, we plot quartiles of (a) neighborhood label entropy and (b) degree, i.e., each bin has 25\% of the samples in sorted order. 
		Overall, we observe that \method{} performs  better than Kipf-GCN and GAT at all levels of node entropy and degree. 
		Please see \refsec{sec:entropy} for details.}
\end{figure*}

\section{Results}
\label{sec:results}

In this section, we attempt to answer the following questions:
\begin{itemize}[itemsep=2pt,topsep=4pt,parsep=0pt,partopsep=0pt]
	\item[Q1.] How does \method{} compare against existing methods for the semi-supervised node classification task? (\refsec{sec:quantitative_results})
	\item[Q2.] How do the performance of methods vary with increasing node degree and neighborhood label mismatch? (\refsec{sec:entropy}) 
	\item[Q3.] How does increasing the number of layers effect \method{}'s performance? (\refsec{sec:layers_results})
	\item[Q4.] What is the effect of ablating different terms in \method{}'s loss function? (\refsec{sec:ablation_results})
	
\end{itemize}

\subsection{Node Classification}
\label{sec:quantitative_results}

The evaluation results for semi-supervised node classification are summarized in \reftbl{tbl:qualitative_results}. Results of all other baseline methods on Cora, Citeseer and Pubmed datasets are taken from \citep{Liao2018graph,graph_attention_network} directly. For evaluation on the Cora-ML dataset, only top performing baselines from the other three datasets are considered. Overall, we find that \method{} outperforms all existing approaches consistently across all the datasets. 

This may be attributed to \method{}'s ability to model nodes' label distribution along with the confidence scores which subdues the effect of noisy nodes during neighborhood aggregation. The lower performance of GAT \citep{graph_attention_network} compared to Kipf-GCN on Cora-ML shows that computing attention based on the hidden representation of nodes is not much helpful in suppressing noisy neighborhood nodes. 
We also observe that the performance of GPNN \citep{Liao2018graph} suffers on the Cora-ML dataset. This is due to the fact that while propagating information between small subgraphs, the high label mismatch rate in Cora-ML (please see \reftbl{tbl:dataset_statistics}) leads to wrong information propagation. Hence, during the global propagation step, this error is further magnified.

\subsection{Effect of Node Entropy and Degree on Performance}
\label{sec:entropy}
In this section, we provide an analysis of the performance of Kipf-GCN, GAT and \method{} for node classification on the Cora-ML dataset which has higher label mismatch rate. We use neighborhood label entropy  to quantify label mismatch, which for a node $u$ is defined as follows.
\[
	\mathrm{Neighbor Label Entropy}(u) = - \sum_{l=1}^{L} p_{ul}  \log p_{ul}\]
\[
	\text{where,}\,\,\, p_{ul} = \frac{|\{v \in \mathcal{N}(u)~|~\mathrm{label}(v) = l \}|}{|\mathcal{N}(u)|}.
\]

Here, $\mathrm{label}(v)$ is the true label of node $v$. The results for neighborhood label entropy and node degree are summarized in Figures \ref{fig:entropy_plot} and \ref{fig:degree_plot}, respectively. On the x-axis of these figures, we plot quartiles of label entropy and degree, i.e., each bin has 25\% of the instances in sorted order. With increasing neighborhood label entropy, the node classification task  is expected to become more challenging. We indeed see this trend in \reffig{fig:entropy_plot} where performances of all the methods degrade with increasing neighborhood label entropy. However, \method{} performs comparatively better than the existing state-of-art approaches at all levels of node entropy.

In case of node degree also (\reffig{fig:degree_plot}), we find that \method{} performs better than Kipf-GCN and GAT at all quartiles of node degrees. Classifying sparsely connected nodes (first and second bins) is challenging as very little information is present in the node neighborhood. Performance improves with availability of moderate number of neighboring nodes (third bin), but further increase in degree (fourth bin) results in introduction of many potentially noisy neighbors, thereby affecting performance of all the methods. For higher degree nodes, \method{} gives an improvement of around 3\% over GAT and Kipf-GCN. This shows that \method{}, through its use of label confidences, is able to give higher influence score to relevant nodes in the neighborhood during aggregation while reducing importance of the noisy ones.



\begin{figure}[t]
	\centering
	\includegraphics[width=\linewidth]{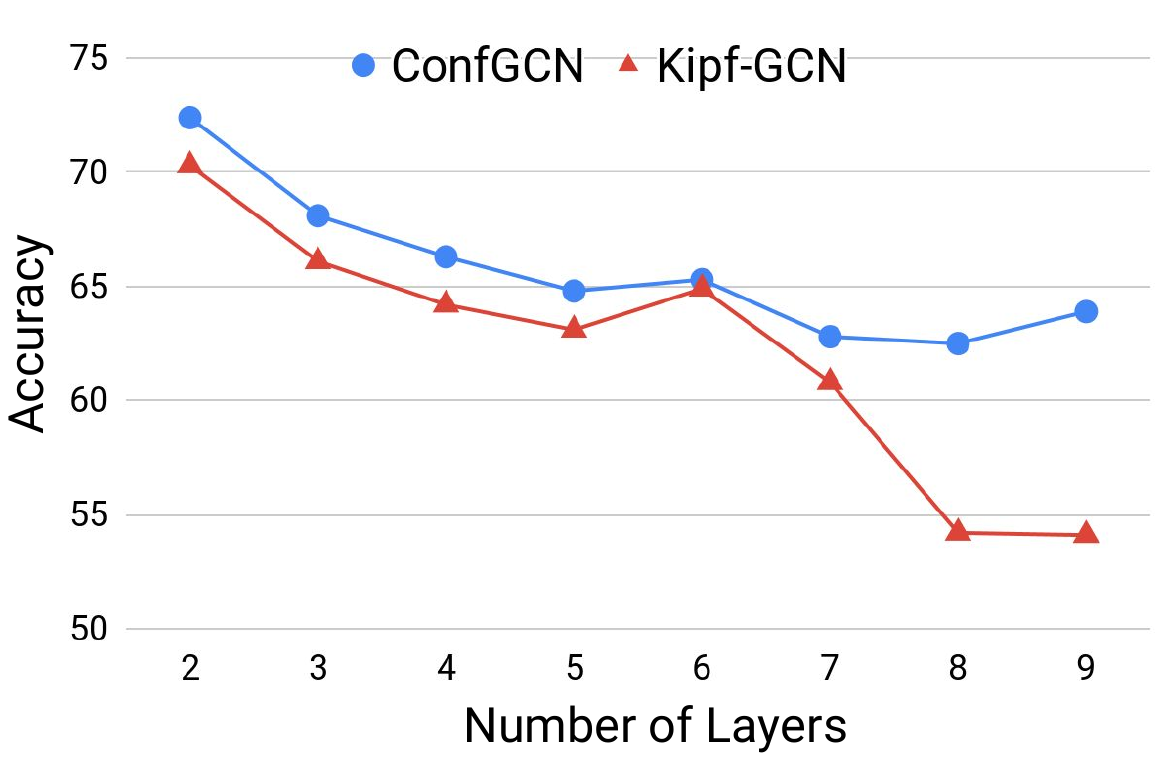}
	\caption{\label{fig:layers} Evaluation of Kipf-GCN and \method{} on the citeseer dataset with increasing number of GCN layers. Overall, \method{} outperforms Kipf-GCN, and while both methods' performance degrade with increasing layers, \method{}'s degradation is more gradual than Kipf-GCN's abrupt drop. 
		Please see \refsec{sec:layers_results} for details.}
\end{figure}

\subsection{Effect of Increasing Convolutional Layers}
\label{sec:layers_results}

Recently, \cite{Xu2018} highlighted an unusual behavior of Kipf-GCN where its performance degrades significantly with increasing number of layers.  
This is because of increase in the number of influencing nodes with increasing layers, resulting in ``averaging out" of information during aggregation.
For comparison, we evaluate the performance of Kipf-GCN and \method{} on citeseer dataset with increasing number of convolutional layers. The results are summarized in \reffig{fig:layers}. We observe that Kipf-GCN's performance degrades drastically with increasing number of layers,  whereas \method{}'s decrease in performance is more gradual. This shows that confidence based GCN helps in alleviating this problem. We also note that \method{} outperforms Kipf-GCN at all layer levels.

\begin{figure}[t]
	\centering
	\includegraphics[width=\linewidth]{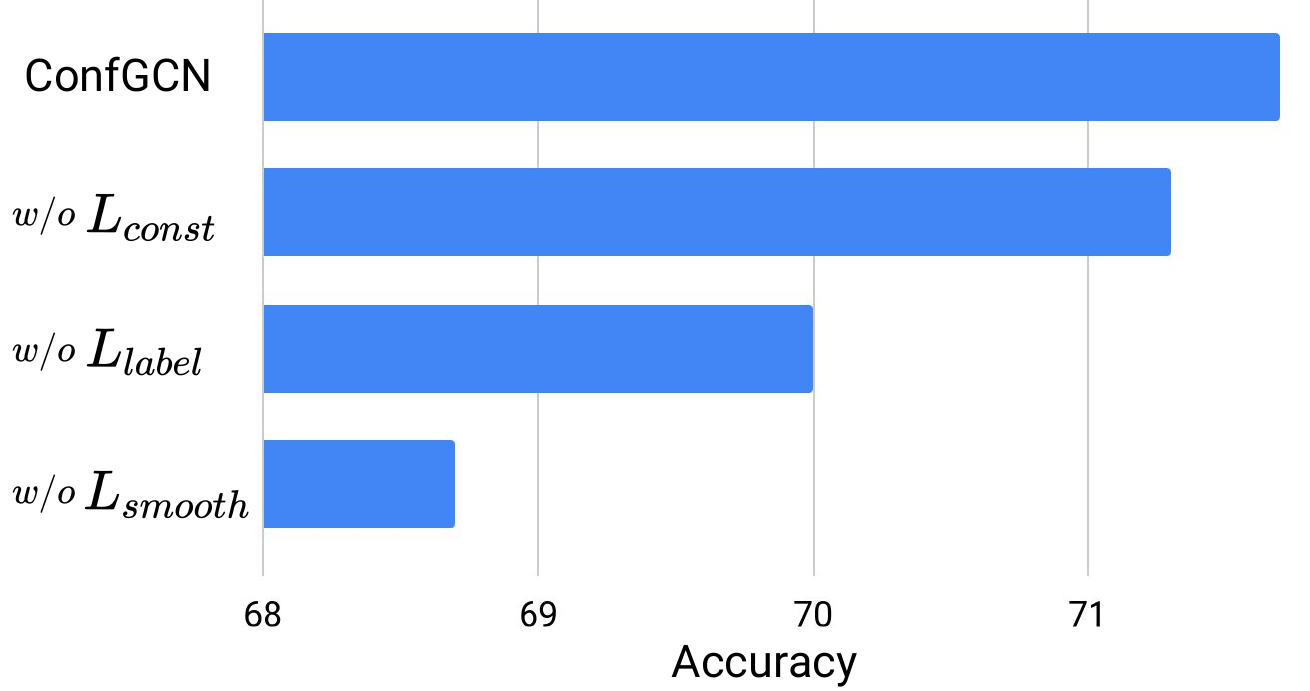}
	\caption{\label{fig:ablation} Performance comparison of different ablated version of \method{} on the citeseer dataset. 
		These results justify inclusion of the different terms in \method{}'s objective function. Please see \refsec{sec:ablation_results} for details.}
\end{figure} 
 
\subsection{Ablation Results}
\label{sec:ablation_results}
In this section, we evaluate the different ablated version of \method{} by cumulatively eliminating terms from its objective function as defined in \refsec{sec:details}. The results on citeseer dataset are summarized in \reffig{fig:ablation}. Overall, we find that each term \method{}'s loss function (\refeqn{eqn:main_obj}) helps in improving its performance and the method performs best when all the terms are included. 


%% file: sections/conclusion.tex

\section{Conclusion}
\label{sec:conclusion}
\vspace{-0.24 cm}

In this paper, we present \method{}, a confidence based Graph Convolutional Network which estimates label scores along with their confidences jointly in a GCN-based setting. In ConfGCN, the influence of one node on another during aggregation is determined using the estimated confidences and label scores, thus inducing anisotropic behavior to GCN. We demonstrate the effectiveness of \method{} against state-of-the-art  methods for the semi-supervised node classification task and analyze its performance in different settings. We make \method{}'s source code available. 
